\documentclass[twoside]{article}

\usepackage[preprint]{aistats2027}

\usepackage[round]{natbib}

\usepackage[utf8]{inputenc}
\usepackage{amsmath,amssymb,amsthm,amsfonts}
\usepackage{bm}
\usepackage{graphicx}
\usepackage{booktabs}
\usepackage{multirow}
\usepackage{microtype}
\usepackage{xcolor}
\usepackage{tikz}
\usetikzlibrary{arrows.meta,positioning,fit,calc,backgrounds}
\usepackage{xurl}
\usepackage[hidelinks]{hyperref}
\usepackage{placeins}

\newtheorem{theorem}{Theorem}
\newtheorem{proposition}[theorem]{Proposition}

\definecolor{resae}{HTML}{1F77B4}
\definecolor{rawtraj}{HTML}{FF7F0E}
\definecolor{pertime}{HTML}{D62728}
\definecolor{shareddict}{HTML}{8C564B}
\definecolor{refgray}{HTML}{7F7F7F}
\definecolor{inkgray}{HTML}{333333}

\begin{document}

\runningauthor{Calvin Yeung, Prathyush Poduval, Ali Zakeri, Zhuowen Zou, Mohsen Imani}

\twocolumn[
\aistatstitle{Residualized Temporal Sparse Autoencoders for Interpreting Diffusion Models}
\aistatsauthor{Calvin Yeung$^{1,*}$ \And Prathyush Poduval$^{1}$ \And Ali Zakeri$^{1}$ \And Zhuowen Zou$^{1}$ \And Mohsen Imani$^{1}$}
\aistatsaddress{$^{1}$Department of Computer Science, University of California, Irvine \\ $^{*}$Corresponding author: \texttt{chyeung2@uci.edu}}
]

\begin{abstract}
Text-to-image diffusion models generate images by iterative denoising, so their internal layers produce trajectories of activations rather than single static representations. Sparse autoencoders (SAEs) have recently been used to decompose diffusion activations into interpretable features, but most approaches analyze individual timesteps or condition on time rather than learning from full trajectories. Training one SAE on whole trajectories would make each feature a single trajectory across timesteps, but adjacent activations are largely linearly predictable from one another, so such an SAE spends its latents on content carried forward from step to step. We introduce residualized temporal SAEs (ReSAE), which fit linear predictors between neighboring timesteps and represent each trajectory by its initial activation and the residuals these dynamics leave unexplained. Training an SAE on this representation is equivalent to training it on raw trajectories under a metric induced by the linear dynamics, and it encourages latents to capture structure beyond what is linearly predictable. Each latent's decoder direction maps back to activation space as a feature trajectory over denoising time. Across Stable Diffusion~1.5 and a Diffusion Transformer, ReSAE features span the whole trajectory while pinpointing when changes enter it, making ReSAE a natural tool for studying how a diffusion model generates an image over time.
\end{abstract}

\section{Introduction}\label{sec:intro}

\begin{figure*}[t]
    \centering
    \includegraphics[width=\textwidth]{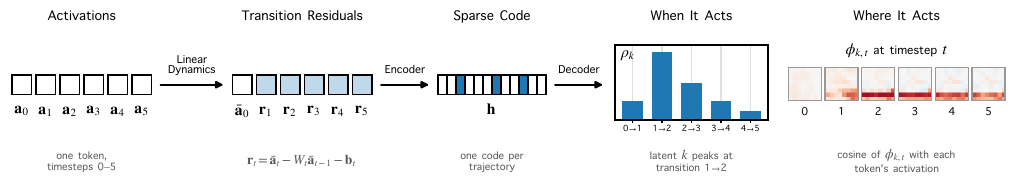}
    \caption{Method Overview. Each activation of a token trajectory after the first is replaced by its transition residual under a linear model of the dynamics, and one sparse dictionary is trained on the whole residualized trajectory. One code per trajectory lets each latent follow one feature across timesteps; mapping its decoder direction back through the linear dynamics gives a feature trajectory, with a temporal profile (when it acts) and a spatial map (where).}
    \label{fig:pipeline}
\end{figure*}

Text-to-image diffusion models generate an image by applying the same denoising network over many timesteps \citep{sohlDicksteinDeepUnsupervised2015,hoDenoisingDiffusionProbabilistic2020,songScoreBasedGenerativeModeling2021}, so an internal layer produces a trajectory of activations rather than a single representation, with coarse structure set early and detail refined late \citep{Choi2022,baranchukLabelEfficientSemantic2022,zhangTaleTwoFeatures2023}. Thus, understanding these activation trajectories is important for interpreting how diffusion models organize and transform visual information over time. Sparse autoencoders (SAEs) decompose activations into sparse, interpretable feature directions \citep{bricken2023monosemanticity,gaoScalingEvaluatingSparse2025}, and in diffusion models such features can be interpretable and causally influence generation \citep{surkovOneStepEnoughSparse2025,tinazEmergenceEvolutionInterpretable2025,shabalinInterpretingLargeText2025}.

However, existing SAEs for diffusion models treat each timestep separately \citep{tinazEmergenceEvolutionInterpretable2025} or share or condition one dictionary across time \citep{surkovOneStepEnoughSparse2025,huangTIDETemporalAware2026}. A natural way to incorporate temporal information is to train an SAE on concatenated activation trajectories, where each latent corresponds to a single feature trajectory across timesteps. Diffusion activations, however, are highly correlated across adjacent timesteps; much of the activation at one timestep is linearly predictable from the activation at a neighboring timestep. Thus, a temporal SAE trained directly on raw activations would spend its latents on content carried forward from one timestep to the next, resulting in sparse features that are spread across time rather than tied to the step where they act. This motivates two desiderata for temporal interpretation: (i) each feature should be a single trajectory across timesteps, and (ii) it should be active mainly at the step where it acts. Achieving both requires learning from whole trajectories while factoring out what each step inherits from the previous one.

Following this principle, we introduce a residualized temporal SAE (ReSAE, Figure~\ref{fig:pipeline}). Rather than training directly on raw activation trajectories, we fit a linear model of the dynamics between neighboring timesteps and replace each activation after the first by its transition residual, i.e., the part that the previous activation does not predict; one SAE is then trained on the residualized trajectory. This encourages sparse features to capture structure beyond linearly predictable information while still preserving their interpretation as feature trajectories over denoising time. Because residualization is an invertible linear map, ReSAE is the trajectory SAE trained under a reconstruction metric induced by the linear dynamics (Section~\ref{sec:metric-view}); this predicts where its explained variance goes and that its decoder directions localize in time. By mapping residualized decoder directions back into activation space, we obtain a feature trajectory for each latent, which lets us measure when a feature acts and where it localizes spatially.

We test both predictions on Stable Diffusion~1.5 \citep{rombachHighResolutionImageSynthesis2022}. Because reconstruction alone cannot show when a feature acts, we also switch the prompt at a known step of generation and test whether that step can be recovered from the few latents that change most.

Our contributions are: (1) ReSAE, a temporal SAE trained on residualized trajectories whose latents map back to feature trajectories; (2) a reconstruction-metric view of ReSAE that predicts its latents localize in time, which we confirm on Stable Diffusion~1.5; (3) a prompt-switch test in which ReSAE latents recover when a change was introduced during generation; and (4) a comparison showing that only ReSAE meets both desiderata, at no cost to steering, and that its localization transfers to a Diffusion Transformer \citep{peeblesScalableDiffusionModels2023}.

\section{Background}\label{sec:background}

\subsection{Text-Conditioned Latent Diffusion Models}\label{sec:diffusion}
Text-to-image latent diffusion models such as Stable Diffusion run a U-Net denoiser $\epsilon_\theta(\mathbf{x}_\tau,\tau,\mathbf{c})$ over denoising steps $\tau$ in a learned latent space, conditioned on a text prompt $\mathbf{c}$ through cross-attention \citep{sohlDicksteinDeepUnsupervised2015,hoDenoisingDiffusionProbabilistic2020,rombachHighResolutionImageSynthesis2022}. With classifier-free guidance (CFG), the U-Net is evaluated both with $\mathbf{c}$ and unconditionally \citep{hoClassifierFreeDiffusionGuidance2022}. 

\subsection{Sparse Autoencoders}\label{sec:sae}
Sparse autoencoders (SAEs) decompose neural network activations into sparse combinations of learned feature directions \citep{bricken2023monosemanticity}. Given an activation vector $\mathbf{x}\in\mathbb{R}^d$, an SAE maps it to a sparse latent representation $\mathbf{h}\in\mathbb{R}^m$ and reconstructs the input,
\begin{align}
    \mathbf{h}=S\!\left(W^{\mathrm{enc}}\mathbf{x}+\mathbf{b}^{\mathrm{enc}}\right),
    \qquad
    \mathbf{x}^{\mathrm{rec}}=W^{\mathrm{dec}}\mathbf{h}+\mathbf{b}^{\mathrm{dec}},
\end{align}
where $S$ is a sparsifying nonlinearity or operator. The columns of $W^{\mathrm{dec}}$ are interpreted as learned feature directions. SAEs are trained to reconstruct activations, $\mathcal{L}_{\mathrm{SAE}}=\|\mathbf{x}-\mathbf{x}^{\mathrm{rec}}\|_2^2$, while keeping the latent code sparse. We use BatchTopK SAEs, which keep the largest latent activations across a minibatch \citep{gaoScalingEvaluatingSparse2025,bussmannBatchTopK2024}. Because an SAE is trained only to reconstruct, the reconstruction loss decides what its dictionary represents; Section~\ref{sec:metric-view} makes this precise for activation trajectories.

\subsection{Sparse Autoencoders for Diffusion Models and for Sequences}\label{sec:related}
SAEs trained on diffusion activations learn interpretable features that causally affect generation and evolve over denoising time \citep{surkovOneStepEnoughSparse2025,tinazEmergenceEvolutionInterpretable2025,shabalinInterpretingLargeText2025}, and SAE features have been used for controllable generation and concept unlearning \citep{kimConceptSteerers2025,cywinskiSAeUron2025}. These approaches handle the denoising trajectory in different ways. \citet{tinazEmergenceEvolutionInterpretable2025} train a separate SAE at each timestep and relate features across timesteps afterwards; \citet{surkovOneStepEnoughSparse2025} reuse a dictionary trained at one step; and \citet{huangTIDETemporalAware2026} share one timestep-conditioned dictionary across timesteps. ReSAE instead learns a single dictionary from whole trajectories, so that each feature spans all timesteps with its own time course; Sections~\ref{sec:design} and~\ref{sec:dit} compare ReSAE with per-timestep, shared and timestep-conditioned dictionaries.

In language models, temporal structure has been built into SAEs through a loss that encourages consistent feature activations across neighboring tokens \citep{bhallaTemporalSAE2025}, and through a decomposition of each activation into a component predictable from the context and a novel residual, a decomposition that reveals event boundaries \citep{lubanaPriorsInTime2025}. These methods learn a code for each token, whereas desideratum (i) of Section~\ref{sec:intro} requires one code per denoising trajectory, so that each latent follows one feature across denoising time.

\section{Methods}

\subsection{Collecting Activation Trajectories}\label{sec:collect-activations}
Consider a diffusion model that generates a sample $\mathbf{x}_0$ from Gaussian noise through a sequence of denoising steps. We use $\tau$ to denote a full denoising step in the sampler and extract an activation tensor $\mathbf{a}(\tau)$ from a fixed layer of the denoising network $\epsilon_\theta$. The full activation trajectory is the sequence of these layer activations over the denoising process. On Stable Diffusion~1.5, $\mathbf{a}(\tau)$ is the update (output minus input) of the mid-block cross-attention layer on the conditional branch of classifier-free guidance, where text conditioning gives interpretable SAE features \citep{tinazEmergenceEvolutionInterpretable2025}.

Each collected activation has shape $H\times W\times d$, where $H$ and $W$ are spatial dimensions and $d$ is the channel dimension. We write $q=(n,s)$ for spatial position $s\in\{1,\ldots,S\}$ of generated sample $n$, and call the sequence of activations at one position a token trajectory. We subsample the denoising trajectory every $\ell$ sampler steps and use $i=0,\ldots,T-1$ for the resulting subsampled timestep indices. If $\tau_i$ is the full sampler step associated with subsampled index $i$, then $\mathbf{a}_{q,i}\in\mathbb{R}^d$ denotes the activation for token trajectory $q$ at full step $\tau_i$. With $\ell=10$ and a $50$-step sampler, this gives $T=6$ timesteps $t=0,\ldots,5$, at $\tau=981,801,601,401,201$ and $1$, and five transitions $t-1\to t$ between them. Equations use $i$ for the same index. Further details are included in Appendix~\ref{app:implementation-details}.

\subsection{Linear Residualization}\label{sec:linear-residualization}
Activation trajectories in diffusion models are highly correlated across time. We first normalize activations separately at each subsampled timestep using training-set statistics, $\bar{\mathbf{a}}_{q,i}=(\mathbf{a}_{q,i}-\boldsymbol{\mu}^{a}_i)\odot(\boldsymbol{\sigma}^{a}_i)^{-1}$, where $\boldsymbol{\mu}^{a}_i$ and $\boldsymbol{\sigma}^{a}_i$ are the per-coordinate mean and standard deviation for activation block $i$, shared across spatial positions, so that high-variance coordinates do not dominate the regression below.

We then remove the linearly predictable component between consecutive normalized activation blocks. For each adjacent subsampled transition $i-1\rightarrow i$, $i=1,\ldots,T-1$, and each spatial position $s$, we fit a ridge regressor $(W_{s,i},\mathbf{b}_{s,i})$ across the training samples $n$:
\begin{align}
    \min_{W_{s,i},\mathbf{b}_{s,i}}
    &\sum_n
    \bigl\|
    \bar{\mathbf{a}}_{(n,s),i}
    -
    W_{s,i}\bar{\mathbf{a}}_{(n,s),i-1}-\mathbf{b}_{s,i}
    \bigr\|_2^2
    \nonumber\\
    &+\lambda\|W_{s,i}\|_F^2.
\end{align}
The transition residual is $\mathbf{r}_{q,i}=\bar{\mathbf{a}}_{q,i}-(W_{s,i}\bar{\mathbf{a}}_{q,i-1}+\mathbf{b}_{s,i})$ for $i=1,\ldots,T-1$, with $s$ the position of token trajectory $q$. Thus, $\mathbf{r}_{q,i}$ captures the component of the normalized activation at timestep $i$ that is not linearly predictable from the previous normalized activation in the subsampled trajectory. A simpler choice would be adjacent differencing, i.e., $W_{s,i}=I$ with no bias; however, the fitted predictor removes substantially more of the predictable component (Appendix~\ref{app:implementation-details}).

\subsection{Temporal Sparse Autoencoder}\label{sec:temporal-sae}
For each token trajectory $q$, we construct a residualized trajectory vector
\begin{align}
    \mathbf{z}_q
    =
    \left[
    \bar{\mathbf{a}}_{q,0},
    \mathbf{r}_{q,1},
    \ldots,
    \mathbf{r}_{q,T-1}
    \right]
    \in\mathbb{R}^{Td}.
\end{align}
The first term anchors the trajectory, while the remaining terms describe deviations from the fitted linear dynamics.

Before SAE training, we apply a second normalization to each block separately, because the first activation and the residuals have different scales and one block would otherwise dominate the reconstruction loss. It sets $\mathbf{x}_{q,i}=(\mathbf{z}_{q,i}-\boldsymbol{\mu}^{z}_i)\odot(\boldsymbol{\sigma}^{z}_i)^{-1}$ with training-split statistics, where $\mathbf{z}_{q,0}=\bar{\mathbf{a}}_{q,0}$ and $\mathbf{z}_{q,i}=\mathbf{r}_{q,i}$ for $i\geq1$; the SAE input is $\mathbf{x}_q=[\mathbf{x}_{q,0},\ldots,\mathbf{x}_{q,T-1}]$.

The SAE encodes and reconstructs the normalized residualized trajectory as $\mathbf{h}_q=f_{\mathrm{enc}}(\mathbf{x}_q)$ and $\mathbf{x}^{\mathrm{rec}}_q=f_{\mathrm{dec}}(\mathbf{h}_q)$. We use BatchTopK SAEs, described in Section~\ref{sec:sae}, with average sparsity $K_{\mathrm{avg}}$ active latents per token trajectory. We train with the reconstruction loss
\begin{align}
    \mathcal{L}_{\mathrm{SAE}}
    =
    \sum_{q\in\mathcal{B}}
    \left\|
    \mathbf{x}_q-\mathbf{x}^{\mathrm{rec}}_q
    \right\|_2^2+\beta\mathcal{L}_\text{aux},
\end{align}
where $\mathcal{L}_\text{aux}$ is the auxiliary loss for dead-latent recovery as in \citet{bussmannBatchTopK2024}. Refer to Appendix~\ref{app:implementation-details} for specific implementation details.

\subsection{Residualization as a Change of Metric}\label{sec:metric-view}
Residualization is an invertible linear transformation of the normalized trajectory, so training on residualized trajectories is the same as training on raw trajectories under a different reconstruction metric. Write $\bar{\mathbf{a}}_q=[\bar{\mathbf{a}}_{q,0},\ldots,\bar{\mathbf{a}}_{q,T-1}]\in\mathbb{R}^{Td}$ and let $L_s$ be the block lower-bidiagonal map with identity diagonal blocks and $-W_{s,i}$ on the subdiagonal, so that $\mathbf{z}_q=L_s\bar{\mathbf{a}}_q-\mathbf{c}_s$ with $\mathbf{c}_s$ collecting the biases. Let $\Sigma=\mathrm{diag}(\boldsymbol{\sigma}^z)$ be the second normalization.

\begin{proposition}\label{prop:metric}
Every reconstruction $\mathbf{x}^{\mathrm{rec}}_q$ determines a reconstructed trajectory $\hat{\mathbf{a}}_q=L_s^{-1}(\Sigma\mathbf{x}^{\mathrm{rec}}_q+\boldsymbol{\mu}^z+\mathbf{c}_s)$, and
\begin{align}
    \left\|\mathbf{x}_q-\mathbf{x}^{\mathrm{rec}}_q\right\|_2^2
    &=
    (\bar{\mathbf{a}}_q-\hat{\mathbf{a}}_q)^\top M_s\,(\bar{\mathbf{a}}_q-\hat{\mathbf{a}}_q),
    \nonumber\\
    M_s&=L_s^\top\Sigma^{-2}L_s .
    \label{eq:metric}
\end{align}
\end{proposition}
The proof is a change of variables (Appendix~\ref{app:proof}). An SAE trained on raw concatenated trajectories (NoResid+Concat, Section~\ref{sec:model-variants}) minimizes the same loss with $L_s=I$, a diagonal metric. ReSAE minimizes it under $M_s$, which measures the error in the transition residuals rather than in the activations themselves.

Two predictions follow: \emph{(P1)} Each model is optimized for its own metric, so at matched sparsity ReSAE should reconstruct the raw activations less accurately and the transition residuals more accurately than a raw concatenated SAE. \emph{(P2)} Consider content that enters at transition $i$ and is then carried forward by the dynamics. In raw coordinates it appears at every timestep from $i$ onward, so a latent of a raw concatenated SAE that explains it must place mass on all of those blocks. In residual coordinates it appears only in block $i$, because the linear dynamics predict it afterwards, so a single latent confined to that block explains it at no cost elsewhere. ReSAE's decoder directions should thus concentrate at single transitions. Section~\ref{sec:capacity} tests both predictions.

\subsection{Activation-Space Decoder Directions}\label{sec:decoder-directions}
For residualized temporal SAEs, decoder directions initially live in normalized SAE input space. Undoing the SAE-input normalization of the decoder vector $\mathbf{d}_k$ of latent $k$ gives $\boldsymbol{\sigma}^{z}\odot\mathbf{d}_k=[\mathbf{v}_{k,0},\mathbf{u}_{k,1},\ldots,\mathbf{u}_{k,T-1}]$, and mapping it back through the linear dynamics of position $s$ gives a decoder trajectory
\begin{align}
    \bar{\boldsymbol{\phi}}_{k,0}
    =
    \mathbf{v}_{k,0},
    \quad
    \bar{\boldsymbol{\phi}}_{k,i}
    =
    W_{s,i}\bar{\boldsymbol{\phi}}_{k,i-1}
    +
    \mathbf{u}_{k,i},
    \quad i\geq1 .
\end{align}
The bias terms do not appear because decoder vectors represent directions rather than absolute activations. Undoing the activation normalization, $\boldsymbol{\phi}_{k,i}=\boldsymbol{\sigma}^{a}_i\odot\bar{\boldsymbol{\phi}}_{k,i}$, gives the feature trajectory $\{\boldsymbol{\phi}_{k,i}\}_{i=0}^{T-1}$ in activation units, which we use for temporal analysis, spatial localization, and steering. Because the dynamics are fitted per position, each ReSAE latent has one feature trajectory $\boldsymbol{\phi}_{k,i}(s)$ per spatial position $s$, and we drop $s$ where it is not needed. Temporal profiles average over positions (Section~\ref{sec:attribution}), spatial maps compare the trajectory at each position with the activation there, and steering adds at each token the trajectory of its own position (Appendix~\ref{app:steering}). For the raw concatenated SAE, the same construction applies with $W_{s,i}=0$, so its feature trajectory is simply its decoder direction and is the same at every position.

\subsection{Temporal Profiles and Temporal Attribution}\label{sec:attribution}
We summarize when a latent acts by how much its feature trajectory changes at each transition beyond what the linear dynamics predict. For latent $k$ and transition $i$, let $g_{k,i}$ be the root-mean-square norm, over spatial positions, of the standardized transition residual of its decoder trajectory. For ReSAE this residual is simply the decoder block $\mathbf{u}_{k,i}$, which is the same at every position; for other trajectory SAEs we apply $L_s$ to the decoder trajectory and divide by $\boldsymbol{\sigma}^z_i$. Normalizing gives the \emph{temporal profile}
\begin{align}
    \rho_{k,i}=\frac{g_{k,i}}{\sqrt{\tfrac{1}{T-1}\sum_{j=1}^{T-1} g_{k,j}^2}},
    \qquad i=1,\ldots,T-1 ,
\end{align}
so that $\tfrac{1}{T-1}\sum_i\rho_{k,i}^2=1$. We summarize a profile by its \emph{peak ratio} $\max_i\rho_{k,i}^2$, which ranges from $1$, for a latent that acts equally at every transition, to $T-1$, for one that acts at a single transition. A peak ratio of $r$ places a fraction $r/(T-1)$ of the profile's squared mass on one transition.

To check that localization holds without the fitted dynamics, we also compute profiles from adjacent differences $\boldsymbol{\phi}_{k,i}-\boldsymbol{\phi}_{k,i-1}$ (Appendix~\ref{app:attribution}).

Temporal profiles also let us attribute a change between two generations to the transitions at which it entered. We take the few latents whose codes change most and add up their profiles, each weighted by how much its code changed. Formally, let $\Delta\bar h_k=\tfrac{1}{|\mathcal{R}|}\sum_{q\in\mathcal{R}}(h'_{q,k}-h_{q,k})$ be the change in the code of latent $k$ averaged over a spatial region $\mathcal{R}$, and $\mathcal{S}$ a small set of latents with the largest $|\Delta\bar h_k|$. The \emph{attribution profile} is
\begin{align}
    E_i=\sum_{k\in\mathcal{S}}\left|\Delta\bar h_k\right|\rho_{k,i},
    \qquad i=1,\ldots,T-1 .
    \label{eq:attribution}
\end{align}
Given the set $\mathcal{C}$ of transitions at which a change can occur, the \emph{attributed onset} is the first $i\in\mathcal{C}$ at which $E_i$ reaches half of $\max_{j\in\mathcal{C}}E_j$. Attribution involves no fitted parameters beyond the size of $\mathcal{S}$.

\subsection{Model Variants}\label{sec:model-variants}
We compare variants along two axes: whether the SAE input is the residualized trajectory $[\bar{\mathbf{a}}_{q,0},\mathbf{r}_{q,1},\ldots,\mathbf{r}_{q,T-1}]$ or the raw one $[\bar{\mathbf{a}}_{q,0},\ldots,\bar{\mathbf{a}}_{q,T-1}]$, each block normalized with its own statistics, and whether one SAE is trained on the concatenated trajectory or one per timestep, as in \citet{tinazEmergenceEvolutionInterpretable2025}. This gives ReSAE (Resid+Concat), NoResid+Concat, Resid+NoConcat and NoResid+NoConcat. We also compare a Shared Dictionary, one dictionary trained on all timesteps and applied to each separately, and a Matryoshka SAE (Appendices~\ref{app:sae} and~\ref{app:design}).

\section{Results}\label{sec:results}
We evaluate ReSAE in three steps. We first check that residualization changes what the dictionary learns, as Section~\ref{sec:metric-view} predicts (Sections~\ref{sec:capacity} and~\ref{sec:spatiotemporal}). We then test whether the resulting latents locate a change whose timing we control (Sections~\ref{sec:onset} and~\ref{sec:attribution-results}). Finally, we compare ReSAE with other temporal SAE designs and apply it to a second architecture (Sections~\ref{sec:design} and~\ref{sec:dit}).

\subsection{Experimental Setup}\label{sec:setup}
Following prior work on sparse autoencoders for diffusion models \citep{surkovOneStepEnoughSparse2025,tinazEmergenceEvolutionInterpretable2025}, we generate 50{,}000 training images and 2{,}500 validation images from LAION-COCO-aesthetic captions \citep{schuhmannLAIONCOCO2022} using Stable Diffusion~1.5. We collect U-Net activation trajectories according to the procedure described in Section~\ref{sec:collect-activations} and evaluate the model variants defined in Section~\ref{sec:model-variants}. Unless stated otherwise, models use $K_{\mathrm{avg}}=50$ active latents per token trajectory and $3{,}840$ latents, and comparisons are at matched sparsity (Appendix~\ref{app:implementation-details}). We measure reconstruction from the sparse code: each model encodes a validation trajectory and decodes it without access to the true activations at other timesteps (Appendix~\ref{app:metrics}). Intervals are 95\% bootstrap intervals over validation images or prompt pairs.

\begin{figure}[t]
    \centering
    \includegraphics[width=\linewidth]{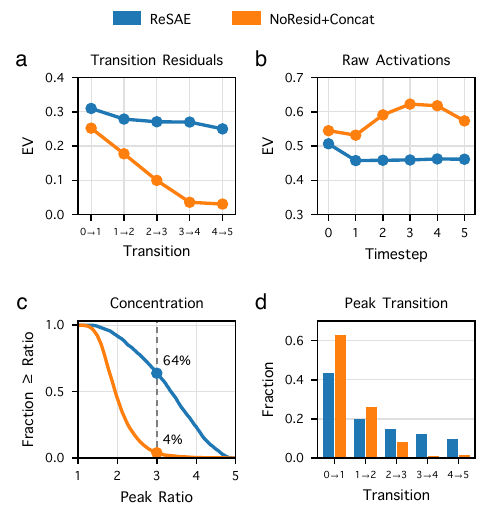}
    \caption{What residualization changes. (a) Explained variance (EV) of the transition residuals at each transition, reconstructed from the sparse code; this evaluates ReSAE's training objective on both models' reconstructions. (b) EV of the raw activations at each timestep, the trade-off the change of metric predicts. (c) Fraction of the $3{,}840$ latents whose temporal profile has at least a given peak ratio ($59\%$ against $37\%$ at a ratio of $3$ when profiles are computed from adjacent differences, without the linear dynamics; Appendix~\ref{app:attribution}). (d) Fraction of latents by the transition at which their profile peaks.}
    \label{fig:mechanism}
\end{figure}

\subsection{Reconstruction Performance}\label{sec:capacity}
Section~\ref{sec:metric-view} predicts that ReSAE trades fidelity to the raw activations for fidelity to the transition residuals (P1), and that its decoder directions concentrate at single transitions (P2). P1 checks that ReSAE spends its latents on what changes rather than on what is carried forward, and P2 checks that each latent is tied to the transition where it acts, as desideratum (ii) of Section~\ref{sec:intro} requires. We test both against NoResid+Concat at matched sparsity, with explained variance (EV) of the raw activations and of the transition residuals (Appendix~\ref{app:metrics}).

\begin{figure*}[t]
    \centering
    \includegraphics[width=\textwidth]{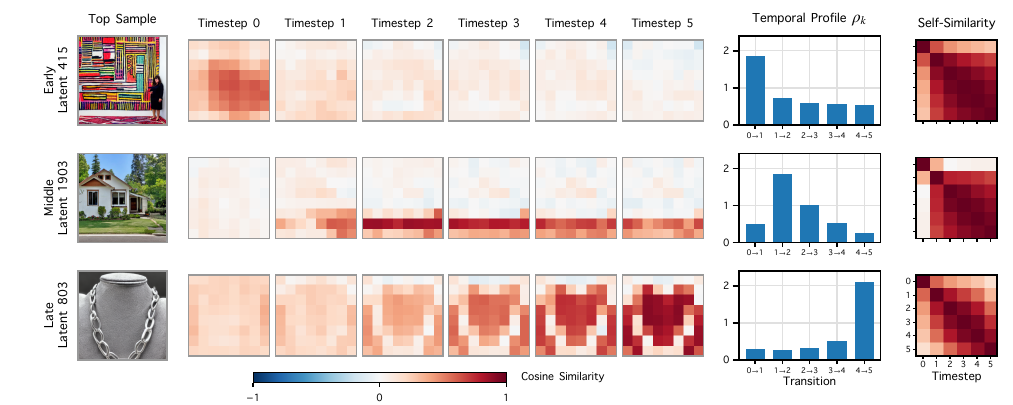}
    \caption{Representative early, middle, and late latents. Left: the latent's top sample. Middle: cosine similarity between the latent's feature trajectory $\boldsymbol{\phi}_{k,t}(s)$ and the sample's activation at each token $s$ and timestep $t$. Right: temporal profile $\rho_k$, and the cosine similarity of the feature trajectory between timesteps, averaged over positions. Each map is strongest around the transition at which the latent's profile peaks. After that transition the self-similarity stays high partly because the linear dynamics carry the direction forward.}
    \label{fig:spatiotemporal-latents}
\end{figure*}

\paragraph{Raw activations and transition residuals.}
Residualization moves explained variance from the raw activations to the transition residuals, as P1 predicts (Figure~\ref{fig:mechanism}a,b). ReSAE explains $0.31$ of the variance of the transition residuals against $0.18$ for NoResid+Concat, and the gap widens over denoising, with NoResid+Concat's fidelity to the residuals falling almost to zero at the last transitions while ReSAE's stays nearly constant. NoResid+Concat in turn explains more of the raw activations ($0.58$ against $0.47$), most of all at later timesteps, which it reconstructs largely through content carried forward from the previous timestep. The ordering holds at every sparsity we consider (Appendix~\ref{app:metrics}).

Fidelity to the raw activations depends on how a reconstruction is evaluated. In a trajectory-level view, the whole trajectory is rebuilt from the code, so an error at one transition carries into every later timestep. In a step-level view, each timestep is rebuilt from the code and the true previous timestep, with the same linear dynamics for both models. There the dynamics alone explain $0.67$ of the variance of later timesteps, ReSAE's code raises this to $0.76$ and NoResid+Concat's to $0.72$ ($+0.042$ $[+0.041,+0.042]$), so ReSAE's code better describes what changes at each step and gives up raw fidelity only in the trajectory-level view.

\paragraph{Temporal localization of decoder directions.}
As P2 predicts, most of ReSAE's latents act mainly at one transition. $64\%$ of them place at least three fifths of their temporal profile on a single transition, against $4\%$ of NoResid+Concat's (Figure~\ref{fig:mechanism}c), and the ordering holds when the profile is computed without the linear dynamics (Appendix~\ref{app:attribution}). A ReSAE latent's time course thus says when it acts, while most NoResid+Concat latents spread over the whole trajectory.

\subsection{Spatiotemporal Structure of Learned Latents}\label{sec:spatiotemporal}
We visualize individual ReSAE latents as feature trajectories. Figure~\ref{fig:spatiotemporal-latents} shows representative early, middle, and late latents of ReSAE, chosen by a fixed rule (Appendix~\ref{app:feature-analysis}), with their spatial cosine-similarity maps and temporal profiles. The early latent's map is broad and centered at the first timestep. The middle latent appears at transition $1\to2$ and persists, covering the grassy area in the lower part of its top sample. The late latent's top samples are all necklaces on display busts (Figure~\ref{fig:app-top-samples}); its map grows over the last timesteps and is strongest at the last, over the necklace and the bust.

Beyond these examples, ReSAE has latents for every stage of denoising. Of its latents, $21\%$ peak at the last two transitions, against $3\%$ of NoResid+Concat's, most of which peak at the first transition (Figure~\ref{fig:mechanism}d). NoResid+Concat therefore has few latents that could register a change made late in generation, which the next two sections test directly.

\subsection{Onset Localization from the Sparse Code}\label{sec:onset}
If ReSAE's latents act mainly at one transition, its code should record when a change entered generation. We test this by introducing a change at a known time. For $290$ source--target prompt pairs from RIEBench \citep{surkovOneStepEnoughSparse2025}, we switch the conditioning from the source to the target prompt at one of four known steps. We encode the switched and the source generation and average the difference of their codes over the edited region (the tokens of the edited object) to obtain $\Delta\bar{\mathbf{h}}$. We then train a linear classifier for each model to recover the switch timestep $t^*$ from it (Appendix~\ref{app:onset}). As a reference for the timing information in the activations, we train the same classifier on the activation difference itself, averaged over the same region.

\begin{figure}[t]
    \centering
    \includegraphics[width=\linewidth]{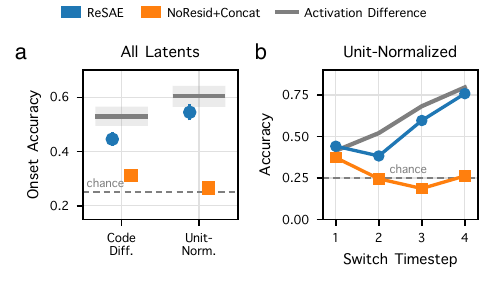}
    \caption{Onset localization from the code difference $\Delta\bar{\mathbf{h}}$ ($1{,}160$ examples, chance $0.25$). A switch at timestep $t^*$ enters at transition $t^*{-}1\to t^*$. (a) Accuracy of a linear classifier on all latents, from $\Delta\bar{\mathbf{h}}$ and after dividing it by its norm (unit-normalized), with 95\% intervals; gray is the same classifier on the activation difference itself, in residual coordinates. (b) Unit-normalized accuracy at each switch timestep.}
    \label{fig:onset}
\end{figure}

\paragraph{ReSAE's code records when the switch occurred.}
ReSAE's code locates the switch with accuracy $0.45$ $[0.42,0.47]$, against $0.31$ $[0.30,0.33]$ for NoResid+Concat (chance $0.25$; Figure~\ref{fig:onset}a). ReSAE keeps its lead when the classifier sees only the five latents most informative about the switch, chosen within each training split ($0.40$ against $0.33$, $+0.064$ $[+0.033,+0.095]$).

\paragraph{ReSAE encodes timing in which latents change.}
Earlier switches change the image more, so a code may reveal the switch through the size of its change alone. To separate the two, we divide $\Delta\bar{\mathbf{h}}$ by its norm. NoResid+Concat's accuracy then falls to chance ($0.27$), while ReSAE's rises to $0.54$. The activation difference itself reaches $0.60$ after the same normalization, so ReSAE's sparse code keeps most of the timing information in the activations.

\paragraph{The advantage comes from the training objective.}
ReSAE's code is a code of residuals, so we test whether its advantage holds in shared coordinates. When both codes are decoded and their differences are expressed in the same residual coordinates, ReSAE keeps its lead ($+0.10$ $[+0.07,+0.13]$; Appendix~\ref{app:onset}). ReSAE is also more accurate at every sparsity and dictionary size we tried (Appendix~\ref{app:onset}).

\begin{figure}[t]
    \centering
    \includegraphics[width=\linewidth]{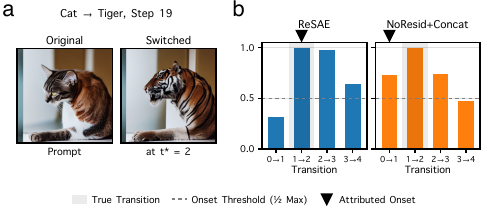}
    \caption{Temporal attribution of a prompt switch. (a) Source image and the image with the prompt switched at $t^*=2$ (cat to tiger). (b) Attribution profile $E_i$ of Eq.~\eqref{eq:attribution}, normalized to its maximum. The example was selected to illustrate a correct ReSAE attribution (Appendix~\ref{app:attribution}).}
    \label{fig:attribution}
\end{figure}

\subsection{Temporal Attribution}\label{sec:attribution-results}
Beyond decoding the onset with a classifier, we test whether it can be read directly from the temporal profiles of a few latents. We attribute each switch to the latents whose codes change most and read the onset from their temporal profiles alone, with no fitted parameters (Eq.~\eqref{eq:attribution}; Figure~\ref{fig:attribution}). We use five latents, the number fixed in advance for the onset classifier of Section~\ref{sec:onset}. ReSAE's attributed onset equals the true onset in $0.60$ $[0.57,0.63]$ of the $1{,}160$ examples, against $0.25$ for NoResid+Concat (chance $0.25$), and it is above chance at every onset (Figure~\ref{fig:app-attribution-confusion}).

NoResid+Concat attributes $98$--$99\%$ of switches at every onset to the first transition, where most of its latents act (Section~\ref{sec:spatiotemporal}), so its attribution carries almost no information about when the change entered, with a Spearman correlation with the true onset of $0.01$, against $0.64$ for ReSAE.

As a control, we generate with the target prompt from the first step, so that the whole difference is present from the start. ReSAE then attributes $260$ of the $290$ generations to the first transition (Appendix~\ref{app:attribution}). Thus, a few of ReSAE's latents are enough to temporally localize most changes.

\subsection{Comparison of Temporal SAE Designs}\label{sec:design}
To test whether other temporal SAE designs meet both desiderata of Section~\ref{sec:intro}, we compare ReSAE at matched sparsity with a per-timestep SAE on the same transition residuals (Resid+NoConcat) and with two dictionaries whose latents span all timesteps, the Shared Dictionary and NoResid+Concat (Table~\ref{tab:designs}; Appendix~\ref{app:design}).

\paragraph{Per-timestep SAEs localize change but cannot follow a feature across timesteps.}
With all latents, the per-timestep SAE locates the switch best ($0.63$), because each of its latents belongs to one timestep, so which latents are active already says when. With the five most informative latents, ReSAE is level with it ($0.40$ against $0.42$, $-0.025$ $[-0.053,+0.002]$). A per-timestep SAE, however, cannot follow a latent across timesteps. Following a latent is what gives a feature trajectory (Figure~\ref{fig:spatiotemporal-latents}) and lets a change be attributed to particular latents (Section~\ref{sec:attribution-results}) without matching features across dictionaries afterwards.

\paragraph{Shared dictionaries follow a feature across timesteps but localize change less well.}
The Shared Dictionary and NoResid+Concat locate the switch with accuracy $0.35$ and $0.31$, against $0.45$ for ReSAE, and with five latents ReSAE leads both ($+0.074$ and $+0.064$). Of the designs compared, only ReSAE meets both desiderata.

\paragraph{ReSAE steers as well as the other designs.}
At matched image distortion, ReSAE's code difference gives the largest steering gain of the designs at default sparsity, although most differences between designs lie within their intervals (Appendix~\ref{app:steering}).

\begin{table}[t]
\caption{Temporal SAE designs at matched sparsity ($50$, $50$, $60$ and $48$ active latents per token trajectory). The first three rows hold by construction. Onset accuracy has chance $0.25$; ``five'' restricts the classifier to five latents (Appendix~\ref{app:design}).}\label{tab:designs}
\centering
\scriptsize
\setlength{\tabcolsep}{2.5pt}
\begin{tabular}{@{}lcccc@{}}
\toprule
 & ReSAE & \shortstack{NoResid\\+Concat} & \shortstack{Resid\\+NoConcat} & \shortstack{Shared\\Dict.} \\
\midrule
latent spans all timesteps & yes & yes & no & yes \\
one code per trajectory & yes & yes & no & no \\
decoder varies over time & yes & yes & no & no \\
\midrule
onset acc., all latents & 0.45 & 0.31 & 0.63 & 0.35 \\
onset acc., five latents & 0.40 & 0.33 & 0.42 & 0.32 \\
EV, raw activations & 0.47 & 0.58 & 0.44 & 0.53 \\
EV, transition residuals & 0.31 & 0.18 & 0.33 & 0.20 \\
\bottomrule
\end{tabular}
\end{table}

\subsection{Transfer to a Diffusion Transformer}\label{sec:dit}
ReSAE needs only activations along the trajectory and a linear fit between timesteps, so it applies beyond the U-Net. We apply it to DiT-XL/2 \citep{peeblesScalableDiffusionModels2023}, a class-conditional Diffusion Transformer, with linear dynamics shared across patches, and compare it with a TIDE-style SAE \citep{huangTIDETemporalAware2026}, a single dictionary shared across timesteps whose encoder is conditioned on the timestep, at matched sparsity and parameter count. The two reconstruct the activations about equally well (Appendix~\ref{app:dit}). In place of the prompt switch, we switch the ImageNet class label at one of three known steps (chance $1/3$).

In the comparison we specified in advance, with the five most informative latents, ReSAE locates the switch with accuracy $0.62$ against $0.54$ for the TIDE-style SAE ($+0.08$ $[+0.05,+0.11]$). With all latents, ReSAE reaches $0.80$ against $0.51$, above the $0.68$ of the activation difference itself, and the gap remains after the size of the code change is removed. On this task the size of the code change carries much of the timing, and with five latents and that size removed, the two dictionaries are level (Appendix~\ref{app:dit}). ReSAE thus carries over to a transformer and again locates the switch better than a timestep-conditioned shared dictionary.

\section{Discussion}\label{sec:discussion}
\paragraph{The reconstruction metric determines what learned features describe.}
With architecture and sparsity fixed, the metric decides whether a code describes what is present at each timestep or what changes at each transition. Designing a sparse dictionary therefore means choosing the metric that fits the question, and reconstruction fidelity alone cannot rank dictionaries for interpretation. Proposition~\ref{prop:metric} gives this choice a statistical reading, since $M_s$ is the precision matrix of a first-order linear-Gaussian model of the trajectory. ReSAE thus scores a reconstruction by its Mahalanobis distance under the fitted dynamics, while the diagonal metric of the raw concatenated SAE treats timesteps as independent (Appendix~\ref{app:proof}). For diffusion models, a metric matched to the dynamics turns a dictionary into a record of when each feature enters generation, which is the question a temporal account of image formation has to answer.

\paragraph{Sparse dictionaries can decompose a process.}
Sparse dictionaries are usually trained on a network's states, so their features describe what is represented at one point of a computation. Training on what a model of the dynamics leaves unexplained instead yields features that enter at one step and are then carried forward; a computation over time becomes a sparse set of such changes, only requiring activations along a trajectory and a model of how they evolve. Sparse dictionaries can thus describe how a network builds its output over time, in addition to what it represents.

\paragraph{Limitations.}\label{sec:limitations}
We train one model per variant and study one layer of each model under one sampling schedule. The dynamics are linear, so content that is only nonlinearly predictable still reaches the latents. Attribution shows when and where a latent changes; its meaning and causal role remain open, and five latents do not identify the kind of edit (Appendix~\ref{app:onset}).

\section{Conclusion}\label{sec:conclusion}
We introduced residualized temporal SAEs, which train one sparse dictionary on what linear dynamics between timesteps leave unexplained, so that each latent follows one feature across the trajectory and acts mainly at one transition. On Stable Diffusion~1.5, the few latents that change most reveal when a prompt change entered, and of the designs compared only ReSAE meets both desiderata. On a Diffusion Transformer, ReSAE again locates a change better than a timestep-conditioned dictionary. Choosing what a dictionary reconstructs thus lets sparse features describe both what a diffusion model represents and when it enters generation.

\subsubsection*{Acknowledgments}
This work was supported in part by the DARPA Young Faculty Award, the National Science Foundation (NSF) under Grants \#2127780, \#2319198, \#2321840, \#2312517, and \#2235472, \#2431561, the Semiconductor Research Corporation (SRC), the Office of Naval Research through the Young Investigator Program Award, and Grants \#N00014-21-1-2225 and \#N00014-22-1-2067, Army Research Office Grant \#W911NF2410360. Additionally, support was provided by the Air Force Office of Scientific Research under Award \#FA9550-22-1-0253, along with generous gifts from Xilinx and Cisco.

\subsection*{AI Use Statement}
In this work, we used generative AI tools to assist with editing and restructuring the manuscript text, with writing plotting code for the figures from exported result files, and with writing and running analysis code for the evaluations. We did not use generative AI tools to generate research data or to produce the reported numbers other than through code that the authors specified and reviewed. We have reviewed all AI-assisted work, and take responsibility for the final content of this work.

\bibliography{main}

\clearpage
\appendix
\let\appendixsection\section
\renewcommand{\section}{\FloatBarrier\appendixsection}
\thispagestyle{empty}
\onecolumn
\aistatstitle{Residualized Temporal Sparse Autoencoders for Interpreting Diffusion Models: \\ Supplementary Materials}
\section{Implementation Details}\label{app:implementation-details}

\subsection{Generation and Dataset}
All experiments use Stable Diffusion 1.5 (\texttt{runwayml/stable-diffusion-v1-5}). Prompts are the captions of \texttt{guangyil/laion-coco-aesthetic} (\texttt{train} split), streamed in an order shuffled with seed 42 and buffer size 10{,}000. The first 50{,}000 captions give the training images and the next 2{,}500 the validation images. Images are generated at $512\times512$ resolution with 50 sampling steps, classifier-free guidance scale 7.5, and an empty negative prompt, using the PNDM sampler. Classifier-free guidance combines the conditional and unconditional predictions as
\begin{align}
    \epsilon_{\mathrm{cfg}}(\mathbf{x}_\tau,\tau,\mathbf{c})
    =
    \epsilon_\theta(\mathbf{x}_\tau,\tau,\varnothing)
    +
    w\left(\epsilon_\theta(\mathbf{x}_\tau,\tau,\mathbf{c})-\epsilon_\theta(\mathbf{x}_\tau,\tau,\varnothing)\right).
\end{align}

\paragraph{Existing assets and licenses.}
Stable Diffusion 1.5 is distributed under the CreativeML OpenRAIL-M license and associated use restrictions. We use only the \texttt{caption} field of \texttt{guangyil/laion-coco-aesthetic} for prompts; this dataset is distributed on Hugging Face under the Apache-2.0 license and is derived from LAION-COCO synthetic captions for publicly available web images. The prompt pairs of the onset-localization task come from RIEBench \citep{surkovOneStepEnoughSparse2025}, which builds on PIE-Bench \citep{juDirectInversion2024}, and the Diffusion Transformer is DiT-XL/2 \citep{peeblesScalableDiffusionModels2023}.

\subsection{Activation Collection}
We collect activations from the conditional branch of classifier-free guidance. We use the update (output minus input) of the cross-attention layer of the U-Net mid-block.

For stride $\ell=10$, each subsampled activation has shape $8\times8\times1280$, corresponding to $S=64$ spatial tokens and channel dimension $d=1280$. We keep the activations at $(\tau_0,\ldots,\tau_5)=(981,801,601,401,201,1)$, giving $T=6$ timesteps and five transitions. Each generated image contributes 64 spatial-token trajectories, so the training dataset contains $50{,}000\times64=3{,}200{,}000$ token trajectories, and the validation dataset contains $2{,}500\times64=160{,}000$ token trajectories.

For residualized models, we fit ridge predictors between neighboring subsampled timestep indices. We fit one ridge predictor per adjacent transition $i-1\rightarrow i$ and per spatial position $s$, across the training images, so that $W$ has shape $5\times64\times1280\times1280$. The ridge penalty is $0.1$. The bias is not penalized. The normalization statistics are per channel and timestep and shared across positions; only the predictors depend on position.

\paragraph{Learned predictor and adjacent differencing.}
On the full validation set, in normalized coordinates, adjacent differencing (the predictor with $W_{s,i}=I$ and no bias) explains $0.574$ $[0.571,0.577]$ of the variance of timesteps $1$--$5$, and the learned predictor explains $0.665$ $[0.662,0.667]$. The learned predictor is better at every transition, reducing the unexplained variance by between $7\%$ and $31\%$, and overall it removes $21.2\%$ $[21.0\%,21.4\%]$ of the variance that differencing leaves.

\paragraph{Temporal stride.}
The cosine self-similarity of the activations over the full 50-step trajectory has a short early block, covering roughly the first $10$ to $15$ steps, followed by a much longer block whose activations differ mainly in scale. A uniform grid needs a stride that divides the 50 steps. Representing each activation of the full trajectory by the most recent retained activation, scaled by the least-squares factor for each image and spatial token, gives mean relative squared errors of $0.048$, $0.142$ and $0.414$ for strides $5$, $10$ and $25$ over $256$ validation images. A stride of $25$ loses much of the temporal variation, and a stride of $5$ nearly doubles the number of timesteps, which sharply increases the size and training cost of a concatenated SAE at a fixed expansion factor. A stride of $10$ keeps timesteps on both sides of the boundary of the early block.

\subsection{SAE Inputs and Training}\label{app:sae}
With six timesteps, the concatenated input of ReSAE and NoResid+Concat has dimension $6\times1280=7{,}680$, and expansion factor $0.5$ gives $3{,}840$ latents. For non-concatenated models, we train one SAE per timestep block, so each timestep-wise SAE has input dimension 1280 and $640$ latents at expansion factor $0.5$.

All main experiments use BatchTopK sparse autoencoders. The SAE uses untied encoder and decoder weights, encoder and decoder biases, a ReLU before BatchTopK selection, Kaiming initialization, and decoder row normalization after each optimizer step. The auxiliary dead-latent loss weight is $1/32$, a latent counts as dead after $2{,}000$ training examples without activating (about eight batches), and the auxiliary top-$k$ value is 256. We train with Adam using learning rate $10^{-4}$, batch size 256, and 30 epochs. Checkpoints are selected by best validation explained variance.

\paragraph{Matched sparsity.}
ReSAE and NoResid+Concat use $K_{\mathrm{avg}}$ active latents per token trajectory. Matching ReSAE's $50$ active latents per trajectory requires about $50/6$ per timestep. In the design comparison of Section~\ref{sec:design}, the Shared Dictionary, a single dictionary of $3{,}840$ latents trained on activations pooled over the six timesteps after per-timestep normalization, uses $8$ per timestep ($48$ per trajectory), and Resid+NoConcat uses $10$ ($60$ per trajectory), slightly more than ReSAE. In the reconstruction comparison of Appendix~\ref{app:metrics}, per-timestep SAEs with $K_{\mathrm{avg}}$ active latents per timestep use $6K_{\mathrm{avg}}$ per trajectory, so Figure~\ref{fig:app-frontier} plots every variant against the number of active latents per token trajectory.

\paragraph{Matryoshka SAE baseline.}
We additionally evaluate a Matryoshka SAE baseline in which latent groups are aligned with temporal chunks of a concatenated activation trajectory. It is trained on the same non-residualized trajectories as NoResid+Concat, with one contiguous latent group $\mathcal{G}_i$ per timestep; each prefix reconstruction $\mathbf{x}^{\mathrm{rec}}_{q,j}=\mathbf{b}_{\mathrm{dec}}+\sum_{i=0}^{j}W_{\mathrm{dec},\mathcal{G}_i}\mathbf{h}_{q,\mathcal{G}_i}$ is trained to reconstruct the full trajectory, the objective averages the error over all prefixes, and sparsity is enforced globally across groups. It is trained with AdamW with weight decay $0.01$, with the other settings unchanged.

\subsection{Compute Resources}\label{app:compute}
Image generation, the linear predictors and SAE training for Stable Diffusion ran on 8 NVIDIA A100 80GB GPUs. Table~\ref{tab:app-cost} separates the costs that any temporal SAE on the same trajectories incurs from those that residualization adds. Residualization adds the linear predictors and one preprocessing pass; the stored activations, the number of timesteps, the input dimension, the dictionary size, the sparsity and the SAE architecture are those of NoResid+Concat. For $M$ token trajectories with $T$ timesteps of width $d$ and a dictionary of $m$ latents, activation storage scales as $O(MTd)$, and the SAE's parameters and per-example computation as $O(Tdm)$, which is $O(T^2d^2)$ at a fixed expansion factor. The per-position predictors add $(T-1)S$ maps of size $d\times d$, fitted once before SAE training. $T$ counts the retained timesteps rather than the sampler steps, so a sampler with more steps does not by itself raise these costs.

\begin{table}[h]
\caption{Compute and storage for the Stable Diffusion experiments ($50{,}000$ training and $2{,}500$ validation images).}\label{tab:app-cost}
\centering\small
\begin{tabular}{@{}lll@{}}
\toprule
item & cost & incurred by \\
\midrule
generating images and extracting activations & about 11 A100 GPU-hours & any temporal SAE \\
activations at all 50 steps & about 409 GiB & not needed for training \\
activations at the six timesteps & about 48 GiB & any temporal SAE \\
SAE inputs & about 92 GiB per model & any temporal SAE \\
fitting the linear predictors and residualizing & under 1 A100 GPU-hour & residualization only \\
training all reported SAE configurations & about 672 A100 GPU-hours & any temporal SAE \\
\bottomrule
\end{tabular}
\end{table}
The other experiments ran on 8 NVIDIA H200 NVL GPUs: onset localization took about 1 GPU-hour, temporal attribution under 0.1, the DiT-XL/2 experiments about 5, and matched-distortion steering about 7.

\section{Reconstruction Metrics}\label{app:metrics}

\paragraph{Trajectory-level reconstruction.}
Each model encodes the true normalized trajectory and rebuilds the whole trajectory from its code. For ReSAE, the decoded first block and residual blocks are mapped back through the linear dynamics, $\hat{\bar{\mathbf{a}}}_{q,0}=\hat{\mathbf{z}}_{q,0}$ and $\hat{\bar{\mathbf{a}}}_{q,i}=W_{s,i}\hat{\bar{\mathbf{a}}}_{q,i-1}+\mathbf{b}_{s,i}+\hat{\mathbf{r}}_{q,i}$, so that an error at one transition propagates to the later timesteps, and the result is unnormalized. For NoResid+Concat, the decoded blocks are unnormalized directly. This is the reconstruction from the sparse code reported throughout.

\paragraph{Step-level reconstruction.}
Each timestep is reconstructed from the code and the true previous timestep, which replaces the reconstructed one in the recursion, $\hat{\bar{\mathbf{a}}}_{q,i}=W_{s,i}\bar{\mathbf{a}}_{q,i-1}+\mathbf{b}_{s,i}+\hat{\mathbf{r}}_{q,i}$. For NoResid+Concat, the implied residual $\tilde{\mathbf{r}}_{q,i}$ is computed from its decoded trajectory with the same linear dynamics and inserted into the same recursion, so that the two models differ only in the residual. With a zero residual the recursion is the linear dynamics alone.

\paragraph{Explained variance.}
The EV of the raw activations uses the Euclidean metric, under which the error at timestep $i$ is $\mathbf{e}_{q,i}=\mathbf{a}_{q,i}-\hat{\mathbf{a}}_{q,i}$ in activation units. The EV of the transition residuals uses the residual metric, under which the error is mapped through $\Sigma^{-1}L_s$ with the same fixed linear dynamics for every model, which by Proposition~\ref{prop:metric} is ReSAE's training error. In both cases
\begin{align}
    \mathrm{EV}=1-\frac{\sum_{i}\mathrm{SSE}_i}{\sum_i\mathrm{SST}_i},
    \qquad
    \mathrm{EV}_i=1-\frac{\mathrm{SSE}_i}{\mathrm{SST}_i},
\end{align}
where $\mathrm{SSE}_i$ sums the squared error at timestep $i$ over validation images, tokens and channels, and $\mathrm{SST}_i$ sums the squared deviation of the true activation from its mean per timestep and channel. Intervals are from $10{,}000$ bootstrap resamples of validation images.

\begin{table}[h]
\caption{Explained variance of reconstruction from the sparse code at $K_{\mathrm{avg}}=50$ and $3{,}840$ latents, with 95\% intervals, and per timestep.}\label{tab:app-ev}
\centering\small
\begin{tabular}{@{}llcl@{}}
\toprule
metric & model & pooled EV & per timestep ($0,\ldots,5$) \\
\midrule
Euclidean & ReSAE & 0.467 [0.464, 0.470] & 0.51, 0.46, 0.46, 0.46, 0.46, 0.46 \\
Euclidean & NoResid+Concat & 0.581 [0.579, 0.584] & 0.54, 0.53, 0.59, 0.62, 0.62, 0.57 \\
residual & ReSAE & 0.311 [0.309, 0.312] & 0.49, 0.31, 0.28, 0.27, 0.27, 0.25 \\
residual & NoResid+Concat & 0.185 [0.183, 0.187] & 0.53, 0.25, 0.18, 0.10, 0.04, 0.03 \\
\bottomrule
\end{tabular}
\end{table}

\begin{figure}[h]
    \centering
    \includegraphics[width=\textwidth]{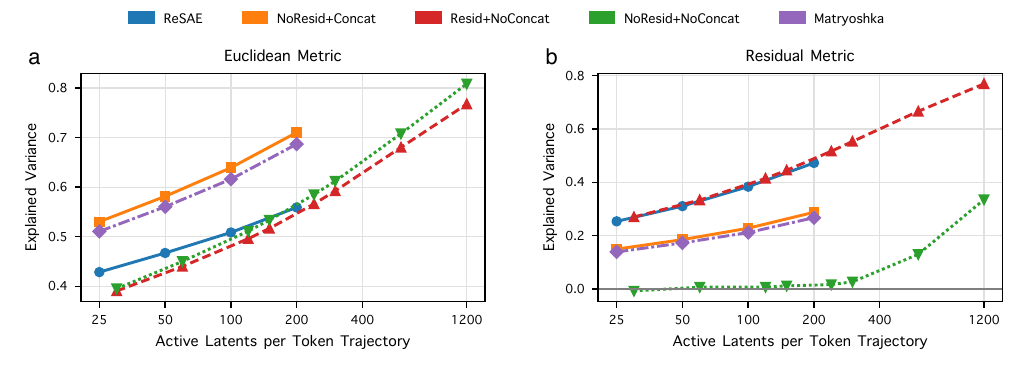}
    \caption{Explained variance of reconstruction from the sparse code as a function of the number of active latents per token trajectory, at expansion factor $0.5$, under the Euclidean metric (a) and the residual metric (b). Per-timestep SAEs use $T=6$ times the per-timestep sparsity.}
    \label{fig:app-frontier}
\end{figure}

The paired differences, ReSAE minus NoResid+Concat, are $-0.114$ $[-0.115,-0.113]$ on the raw activations and $+0.126$ $[+0.124,+0.127]$ on the transition residuals. In the step-level reconstruction, the linear dynamics alone explain $0.666$ of the variance of timesteps $1$--$5$; ReSAE's reconstructed residual raises this to $0.762$, and the residual implied by NoResid+Concat's reconstruction to $0.721$, a paired difference of $+0.042$ $[+0.041,+0.042]$. Figure~\ref{fig:app-frontier} shows explained variance for all variants as a function of the number of active latents per token trajectory. ReSAE explains less of the raw activations and more of the transition residuals than NoResid+Concat at every sparsity shown. At $60$ active latents per trajectory, Resid+NoConcat explains $0.33$ of the transition residuals, close to ReSAE's $0.31$ at $50$, while NoResid+NoConcat explains almost none of them below several hundred active latents.

\section{Proof of Proposition~\ref{prop:metric}}\label{app:proof}
Fix a spatial position $s$ and drop the index $q$. Residualization maps the normalized trajectory to $\mathbf{z}=L_s\bar{\mathbf{a}}-\mathbf{c}_s$, where $L_s$ has identity blocks on the diagonal and $-W_{s,i}$ in block position $(i,i-1)$ for $i=1,\ldots,T-1$, and $\mathbf{c}_s=[\mathbf{0},\mathbf{b}_{s,1},\ldots,\mathbf{b}_{s,T-1}]$. $L_s$ is block lower-triangular with identity diagonal blocks, so $\det L_s=1$ and $L_s$ is invertible. The SAE input is $\mathbf{x}=\Sigma^{-1}(\mathbf{z}-\boldsymbol{\mu}^z)$. For any reconstruction $\mathbf{x}^{\mathrm{rec}}$, define $\hat{\mathbf{z}}=\Sigma\mathbf{x}^{\mathrm{rec}}+\boldsymbol{\mu}^z$ and $\hat{\bar{\mathbf{a}}}=L_s^{-1}(\hat{\mathbf{z}}+\mathbf{c}_s)$, so that $\hat{\mathbf{z}}=L_s\hat{\bar{\mathbf{a}}}-\mathbf{c}_s$. Then
\begin{align}
    \mathbf{x}-\mathbf{x}^{\mathrm{rec}}=\Sigma^{-1}(\mathbf{z}-\hat{\mathbf{z}})=\Sigma^{-1}L_s(\bar{\mathbf{a}}-\hat{\bar{\mathbf{a}}}),
\end{align}
and $\|\mathbf{x}-\mathbf{x}^{\mathrm{rec}}\|_2^2=(\bar{\mathbf{a}}-\hat{\bar{\mathbf{a}}})^\top L_s^\top\Sigma^{-2}L_s(\bar{\mathbf{a}}-\hat{\bar{\mathbf{a}}})$, which is Eq.~\eqref{eq:metric}. $M_s$ is positive definite because $L_s$ is invertible and $\Sigma$ is diagonal with positive entries. Setting $L_s=I$ and $\mathbf{c}_s=\mathbf{0}$ recovers NoResid+Concat, whose loss is the diagonal metric $\Sigma^{-2}$ on the normalized trajectory. $\square$

\paragraph{Likelihood view.}
Suppose that $\mathbf{z}\sim\mathcal{N}(\boldsymbol{\mu}^z,\Sigma^2)$, that is, the first activation and the transition residuals are independent Gaussians with the means and variances of the second normalization. Then $\bar{\mathbf{a}}=L_s^{-1}(\mathbf{z}+\mathbf{c}_s)$ is Gaussian with covariance $L_s^{-1}\Sigma^2L_s^{-\top}$ and precision $L_s^\top\Sigma^{-2}L_s=M_s$, so the loss of Eq.~\eqref{eq:metric} is the squared Mahalanobis distance between the trajectory and its reconstruction under this model. With $L_s=I$ and $\mathbf{c}_s=\mathbf{0}$, the model has independent timesteps, and its precision is the diagonal metric of NoResid+Concat.

\paragraph{Decoder directions under the change of metric.}
Let $\mathbf{d}_k$ be a decoder direction in $\mathbf{x}$-space and $\mathbf{u}=\Sigma\mathbf{d}_k=[\mathbf{u}_0,\ldots,\mathbf{u}_{T-1}]$. Its trajectory in normalized activation space is $\bar{\boldsymbol{\phi}}=L_s^{-1}\mathbf{u}$, which solves $\bar{\boldsymbol{\phi}}_0=\mathbf{u}_0$ and $\bar{\boldsymbol{\phi}}_i=W_{s,i}\bar{\boldsymbol{\phi}}_{i-1}+\mathbf{u}_i$; this is the recursion of Section~\ref{sec:decoder-directions}. If $\mathbf{u}$ is supported on block $j$, then $\bar{\boldsymbol{\phi}}_i=\mathbf{0}$ for $i<j$, $\bar{\boldsymbol{\phi}}_j=\mathbf{u}_j$, and $\bar{\boldsymbol{\phi}}_i=W_{s,i}\cdots W_{s,j+1}\mathbf{u}_j$ for $i>j$. The direction's contribution to the loss is $\|\mathbf{d}_k\|_2^2$ regardless of how far the linear dynamics propagate it, so under $M_s$ a latent explains the residual of one transition and inherits the propagated response at no additional cost. Under the diagonal metric the same trajectory costs $\sum_i\|\Sigma_i^{-1}\bar{\boldsymbol{\phi}}_i\|_2^2$, which includes every propagated timestep. This is prediction (P2).

\section{Onset Localization: Protocol and Additional Results}\label{app:onset}

\paragraph{Prompt switch.}
RIEBench \citep{surkovOneStepEnoughSparse2025} has $296$ source--target prompt pairs, each with an edit category from PIE-Bench \citep{juDirectInversion2024}. We exclude the six pairs whose source or target mask (below) is empty on the $8\times8$ token grid, which leaves $290$. For each pair we generate the source image with the DDIM sampler (50 steps, guidance scale 7.5, empty negative prompt) and, from the same initial noise, four images in which the conditioning changes from the source to the target prompt at step $9$, $19$, $29$ or $39$. Activations are taken at steps $0,9,19,29,39,49$, whose diffusion timesteps $981,801,601,401,201,1$ are those of the training timesteps, so a switch at step $9$, $19$, $29$ or $39$ first changes timestep $1$, $2$, $3$ or $4$ (the switch timestep $t^*$) and enters at transition $0\to1$, $1\to2$, $2\to3$ or $3\to4$. The dictionaries and linear dynamics are applied unchanged, and each dictionary explains as much of the variance of its input on the source and target generations of these pairs as on the validation set ($0.314$ against $0.312$ for ReSAE, $0.577$ against $0.575$ for NoResid+Concat). This gives $1{,}160$ examples. Both generations are encoded with each model, and the per-token code difference is averaged over the tokens of the edited object, given by a Grounded-SAM2 mask \citep{renGroundedSAMAssembling2024} of the source image for the pair's source phrase, with box threshold $0.40$ and text threshold $0.30$ and the whole image if nothing is detected, downsampled to the $8\times8$ token grid by nearest neighbor. The target mask is obtained in the same way from the target image and phrase.

\paragraph{Classifier.}
A multinomial logistic regression predicts the onset from $\Delta\bar{\mathbf{h}}$. We use five outer folds grouped by prompt pair, so that the four examples of a pair are always in the same fold, and choose the inverse regularization strength from $\{10^{-4},\ldots,10^{4}\}$ by inner cross-validation on the training folds. The five-latent classifier keeps the five latents most informative about the switch, those with the largest one-way ANOVA $F$ statistic for the switch timestep, selected within each training split, inner and outer, so that held-out examples play no part in the selection. The unit-normalized classifier divides each $\Delta\bar{\mathbf{h}}$ by its Euclidean norm, and the magnitude classifier sees only that norm. Intervals are from $10{,}000$ bootstrap resamples of prompt pairs. The same protocol is applied to the activation difference without a dictionary, in raw coordinates and after residualization with the fixed linear dynamics, and to both codes after decoding them to activations and residualizing the reconstructions with the same dynamics.

\begin{table}[h]
\caption{Onset accuracy (chance $0.25$) for every representation and classifier, with 95\% intervals for the main comparisons.}\label{tab:app-onset}
\centering\footnotesize
\setlength{\tabcolsep}{3pt}
\resizebox{\textwidth}{!}{\begin{tabular}{@{}lccccc@{}}
\toprule
representation & all latents & five latents & magnitude & all, unit-norm. & five, unit-norm. \\
\midrule
ReSAE code & 0.446 [0.422, 0.472] & 0.396 [0.369, 0.422] & 0.365 & 0.545 [0.515, 0.575] & 0.391 \\
NoResid+Concat code & 0.312 [0.296, 0.328] & 0.332 [0.311, 0.353] & 0.406 & 0.266 [0.244, 0.289] & 0.283 \\
ReSAE, decoded, residual coordinates & 0.442 [0.411, 0.474] & 0.349 & 0.360 & 0.439 & 0.304 \\
NoResid+Concat, decoded, residual coordinates & 0.343 [0.322, 0.365] & 0.340 & 0.400 & 0.312 & 0.286 \\
activation difference, raw & 0.511 [0.480, 0.543] & 0.364 & 0.429 & 0.578 & 0.388 \\
activation difference, residual & 0.529 [0.495, 0.566] & 0.392 & 0.389 & 0.603 & 0.393 \\
\bottomrule
\end{tabular}}
\end{table}

\paragraph{Additional measures.}
Table~\ref{tab:app-onset} lists all accuracies. The mean absolute error in transitions ranks the models the same way ($0.93$ for ReSAE and $1.16$ for NoResid+Concat with all latents, and $0.74$ and $1.23$ after unit normalization). Averaging the code difference over all tokens instead of the edited region gives $0.55$ against $0.32$.

\begin{figure}[h]
    \centering
    \includegraphics[width=\textwidth]{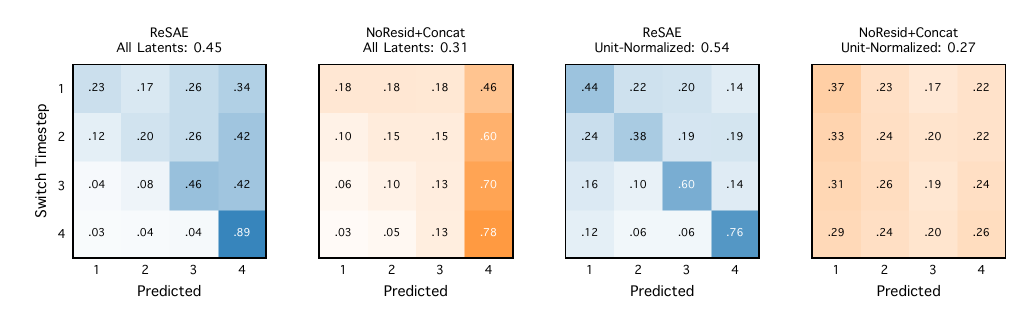}
    \caption{Row-normalized confusion matrices of the onset classifier on all latents, before and after unit normalization of the code difference.}
    \label{fig:app-confusion}
\end{figure}

\paragraph{Magnitude and switch timestep.}
The magnitude $\|\Delta\bar{\mathbf{h}}\|_2$ decreases with the onset in both models (medians $20.3$, $12.3$, $7.8$ and $6.0$ for ReSAE and $18.1$, $8.7$, $4.9$ and $2.9$ for NoResid+Concat), so magnitude alone is informative. Before unit normalization, NoResid+Concat's classifier assigns most examples to the last switch timestep and is right mainly when the switch is late, while ReSAE's also identifies switches at $t^*=3$ ($0.46$ against $0.13$; Figure~\ref{fig:app-confusion}). After unit normalization, ReSAE's accuracy is close to that of the activation difference at every switch timestep except $t^*=2$ (Figure~\ref{fig:onset}b).

\paragraph{Sparsity and dictionary size.}
ReSAE is more accurate than NoResid+Concat at all seven settings of Table~\ref{tab:app-grid}, and the interval of the difference excludes zero in every case except all latents with $15{,}360$ latents ($+0.019$ $[-0.010,+0.047]$).

\begin{figure}[h]
    \centering
    \includegraphics[width=0.5\textwidth]{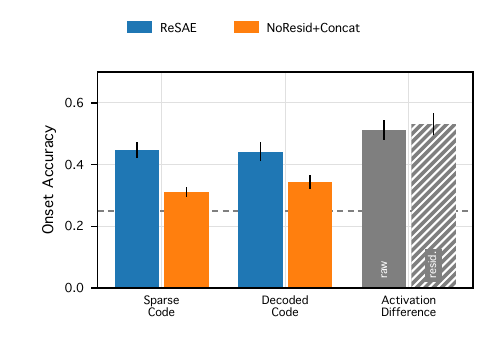}
    \caption{Onset accuracy in common coordinates, with 95\% intervals: the sparse codes, both codes decoded and residualized with the same fixed linear dynamics, and the activation difference itself in raw and residual (hatched) coordinates.}
    \label{fig:app-onset-details}
\end{figure}

\paragraph{Training objective and coordinates.}
ReSAE's code is a code of residuals, so its advantage might reflect the coordinates in which it is expressed. To separate the two, we decode both codes to activations and residualize both reconstructions with the same fixed linear dynamics. ReSAE's reconstruction remains more decodable, $0.44$ against $0.34$, a difference of $+0.10$ $[+0.07,+0.13]$ (Figure~\ref{fig:app-onset-details}), so the advantage comes from the training objective. The activation difference itself decodes the onset about equally well in either coordinate system ($0.51$ and $0.53$), as expected of an invertible linear map.

\paragraph{Edit category.}
To test whether a five-latent summary of the code difference also identifies the kind of edit, we predict each of the nine PIE-Bench edit categories with a one-versus-rest linear classifier on the code difference averaged over all $64$ tokens and restricted to five latents selected by the same rule, at each transition from adjacent differences, with cross-validation folds stratified by category. Macro AUROC is $0.513$ $[0.477,0.550]$ for ReSAE and $0.547$ $[0.511,0.584]$ for NoResid+Concat (difference $-0.034$ $[-0.080,+0.010]$; $0.512$ and $0.549$ in residual coordinates); a coarse three-way grouping into object, attribute and image edits gives $-0.092$ $[-0.152,-0.029]$. The activation difference itself reaches $0.570$ $[0.549,0.590]$, and its category information is higher early than late, from $0.615$ at the first transition to $0.527$ and $0.531$ at the last two. A classifier that sees only the norm of the code difference reaches $0.557$ and $0.573$, so most of what the five-latent summaries carry about the category is magnitude. Category information is present in the activations, more so early than late, and neither five-latent summary captures it.

\begin{table}[h]
\caption{Onset accuracy at seven settings of sparsity and dictionary size, and the paired difference ReSAE minus NoResid+Concat with 95\% intervals.}\label{tab:app-grid}
\centering\small
\setlength{\tabcolsep}{4pt}
\resizebox{\textwidth}{!}{\begin{tabular}{@{}rrcccccc@{}}
\toprule
 & & \multicolumn{3}{c}{all latents} & \multicolumn{3}{c}{five latents} \\
\cmidrule(lr){3-5}\cmidrule(lr){6-8}
$K_{\mathrm{avg}}$ & latents & ReSAE & NoResid+Concat & difference & ReSAE & NoResid+Concat & difference \\
\midrule
25 & 3{,}840 & 0.390 & 0.333 & $+0.057$ [$+0.028$, $+0.087$] & 0.395 & 0.322 & $+0.073$ [$+0.041$, $+0.105$] \\
50 & 1{,}920 & 0.509 & 0.321 & $+0.189$ [$+0.156$, $+0.222$] & 0.441 & 0.331 & $+0.109$ [$+0.077$, $+0.143$] \\
50 & 3{,}840 & 0.446 & 0.312 & $+0.134$ [$+0.104$, $+0.164$] & 0.396 & 0.332 & $+0.064$ [$+0.033$, $+0.095$] \\
50 & 7{,}680 & 0.409 & 0.313 & $+0.097$ [$+0.068$, $+0.126$] & 0.384 & 0.336 & $+0.048$ [$+0.016$, $+0.081$] \\
50 & 15{,}360 & 0.346 & 0.327 & $+0.019$ [$-0.010$, $+0.047$] & 0.386 & 0.314 & $+0.072$ [$+0.040$, $+0.105$] \\
100 & 3{,}840 & 0.505 & 0.325 & $+0.180$ [$+0.150$, $+0.211$] & 0.453 & 0.317 & $+0.135$ [$+0.101$, $+0.169$] \\
200 & 3{,}840 & 0.559 & 0.339 & $+0.220$ [$+0.187$, $+0.253$] & 0.451 & 0.344 & $+0.107$ [$+0.076$, $+0.137$] \\
\bottomrule
\end{tabular}}
\end{table}

\section{Temporal Profiles and Attribution: Details and Examples}\label{app:attribution}

\begin{figure}[h]
    \centering
    \includegraphics[width=\textwidth]{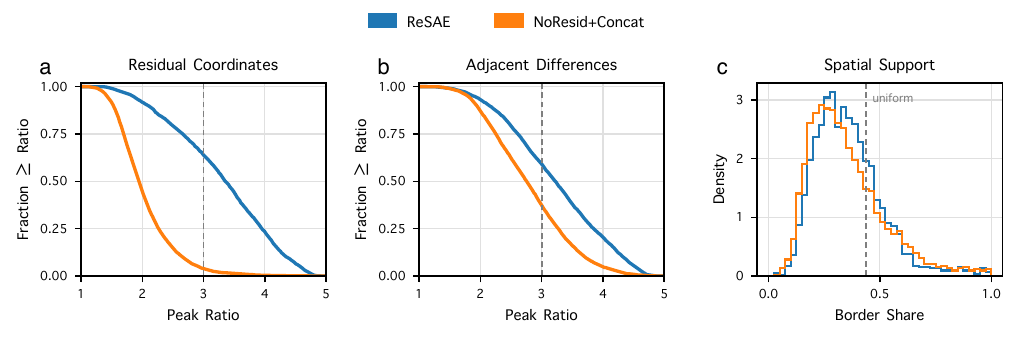}
    \caption{All $3{,}840$ latents of each model. (a,~b) Fraction of latents with at least a given peak ratio, in residual coordinates and from adjacent differences. (c) Distribution of the share of each latent's activation on border tokens.}
    \label{fig:app-localization}
\end{figure}

\paragraph{Coordinates.}
We compute $g_{k,i}$ in two ways. In residual coordinates, used throughout, it is the root-mean-square norm over positions of the standardized transition residual of the decoder trajectory under the fixed linear dynamics, which is the decoder block $\mathbf{u}_{k,i}$ for ReSAE and $L_s$ applied to the decoder trajectory and divided by $\boldsymbol{\sigma}^z_i$ for NoResid+Concat. From adjacent differences, it is the same norm of $\boldsymbol{\phi}_{k,i}-\boldsymbol{\phi}_{k,i-1}$, which does not use the linear dynamics. Figure~\ref{fig:app-localization} shows the distribution of the peak ratio in both. From adjacent differences, $59\%$ of ReSAE's latents and $37\%$ of NoResid+Concat's have a peak ratio of at least $3$, with medians $3.22$ and $2.76$ against $3.36$ and $1.94$ in residual coordinates. In the same coordinates, NoResid+Concat's latents have their largest change at the first transition in $79\%$ of cases, against $45\%$ for ReSAE.

\paragraph{Candidate onsets.}
For the prompt switch the candidate set is $\mathcal{C}=\{1,2,3,4\}$, the transitions at which a switch can enter. Transition $4\to5$ follows every candidate onset, so a latent localized there is consistent with any onset and carries no information about which one occurred. We also report the \emph{attributed peak}, $\arg\max_{i\in\mathcal{C}}E_i$.

\begin{table}[h]
\caption{Attribution accuracy on the prompt switch ($1{,}160$ examples, chance $0.25$), for the five latents with the largest $|\Delta\bar h_k|$ and for the five latents of the onset classifier weighted by their contribution to the predicted class, in both coordinates.}\label{tab:app-attribution}
\centering\small
\setlength{\tabcolsep}{4pt}
\begin{tabular}{@{}llccc@{}}
\toprule
latents & coordinates & model & attributed peak & attributed onset \\
\midrule
largest $|\Delta\bar h_k|$ & residual & ReSAE & 0.537 [0.509, 0.566] & 0.601 [0.570, 0.633] \\
 & & NoResid+Concat & 0.303 [0.285, 0.322] & 0.253 [0.247, 0.258] \\
 & adjacent difference & ReSAE & 0.561 [0.533, 0.590] & 0.459 [0.430, 0.487] \\
 & & NoResid+Concat & 0.265 [0.250, 0.280] & 0.251 [0.250, 0.253] \\
classifier-weighted & residual & ReSAE & 0.359 [0.336, 0.381] & 0.335 [0.314, 0.357] \\
 & & NoResid+Concat & 0.252 [0.241, 0.263] & 0.252 [0.247, 0.257] \\
 & adjacent difference & ReSAE & 0.362 [0.340, 0.384] & 0.333 [0.309, 0.356] \\
 & & NoResid+Concat & 0.262 [0.253, 0.272] & 0.253 [0.247, 0.258] \\
\bottomrule
\end{tabular}
\end{table}

\begin{figure}[h]
    \centering
    \includegraphics[width=0.5\textwidth]{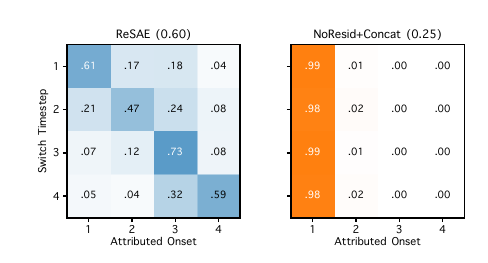}
    \caption{Attributed onset against switch timestep for all $1{,}160$ examples, row-normalized, with each model's accuracy in parentheses.}
    \label{fig:app-attribution-confusion}
\end{figure}

\paragraph{Attribution by switch timestep.}
ReSAE's attributed onset is above chance at every switch timestep (Figure~\ref{fig:app-attribution-confusion}), and its Spearman correlation with the true onset is $0.64$, against $0.01$ for NoResid+Concat, whose attributed onset is transition $0\to1$ in $98$--$99\%$ of examples at every switch timestep. ReSAE's attributed peak agrees with the true onset in $0.54$ of examples against $0.30$ for NoResid+Concat, but it is biased toward later transitions: it is at chance for $t^*=1$ and correct in $0.99$ of examples for $t^*=4$.

\paragraph{Classifier-weighted attribution.}
Weighting the five latents of the onset classifier of Section~\ref{sec:onset} by their contribution to the predicted class, the analysis specified in advance, gives an attributed-peak accuracy of $0.359$ against $0.252$, a difference of $+0.107$ $[+0.084,+0.131]$ (Table~\ref{tab:app-attribution}). The main text uses the five latents with the largest $|\Delta\bar h_k|$, which need no fitted classifier.

\paragraph{Examples.}
The example in Figure~\ref{fig:attribution} was selected to illustrate a correct attribution. For each onset, among the examples whose ReSAE attributed peak and attributed onset both equal the true onset and whose edited region covers 2--50\% of the image, we take the one with the largest masked LPIPS between the source and switched images; NoResid+Concat's attribution played no part in the selection. Figure~\ref{fig:attribution} shows the one selected at $t^*=2$. Among the examples at the 90th percentile of masked LPIPS for each switch timestep, chosen without any condition on the attribution, ReSAE's attributed onset is correct in three of the four and NoResid+Concat's in two. The masked LPIPS falls steeply with the onset (90th percentiles $0.63$, $0.40$, $0.15$ and $0.04$ at $t^*=1,\ldots,4$; Spearman $-0.79$), so switches at later steps change the image little.

\paragraph{Pairs without a switch.}
As a negative control, for the $290$ pairs generated with the target prompt from the first step every change is present from the start, and ReSAE's attributed onset is transition $0\to1$ for $260$ of them.

\section{Comparison of Temporal SAE Designs: Details}\label{app:design}
Resid+NoConcat in Table~\ref{tab:designs} trains one dictionary of $640$ latents on the transition residuals of each timestep (the first timestep on its normalized activation), with $10$ active latents per timestep. The Shared Dictionary trains one dictionary of $3{,}840$ latents on normalized activations pooled over the six timesteps and encodes each timestep separately with $8$ active latents. For the onset classifier, the Resid+NoConcat code of a trajectory is the concatenation of the six per-timestep codes ($3{,}840$ dimensions) and the Shared Dictionary code is the concatenation of the six per-timestep encodings ($23{,}040$ dimensions); for the Shared Dictionary, ``five latents'' means five latents at all six timesteps.

\begin{table}[h]
\caption{Design comparison with 95\% intervals. Differences are ReSAE minus the comparator.}\label{tab:app-design}
\centering\small
\setlength{\tabcolsep}{4pt}
\begin{tabular}{@{}lcccc@{}}
\toprule
 & ReSAE & NoResid+Concat & Resid+NoConcat & Shared Dictionary \\
\midrule
onset, all latents & 0.446 [0.422, 0.472] & 0.312 [0.296, 0.328] & 0.630 [0.601, 0.660] & 0.347 [0.326, 0.369] \\
\quad difference & --- & $+0.134$ [$+0.104$, $+0.164$] & $-0.184$ [$-0.216$, $-0.154$] & $+0.099$ [$+0.070$, $+0.128$] \\
onset, five latents & 0.396 [0.369, 0.422] & 0.332 [0.311, 0.353] & 0.421 [0.402, 0.441] & 0.322 [0.302, 0.342] \\
\quad difference & --- & $+0.064$ [$+0.033$, $+0.095$] & $-0.025$ [$-0.053$, $+0.002$] & $+0.074$ [$+0.041$, $+0.107$] \\
EV, raw activations & 0.467 [0.464, 0.470] & 0.581 [0.579, 0.584] & 0.440 [0.437, 0.443] & 0.526 [0.522, 0.529] \\
EV, transition residuals & 0.311 [0.309, 0.312] & 0.185 [0.183, 0.187] & 0.333 [0.331, 0.335] & 0.202 [0.199, 0.204] \\
\bottomrule
\end{tabular}
\end{table}

\section{Spatiotemporal Structure of Learned Latents}\label{app:feature-analysis}
\paragraph{Selection of the illustrated latents.}
The latents of Figure~\ref{fig:spatiotemporal-latents} are from the ReSAE analyzed throughout ($K_{\mathrm{avg}}=50$, $3{,}840$ latents, six timesteps) and were chosen by a fixed rule. Each latent is scored by its mean encoder activation when active, before the sparsity step, times one minus its activation frequency, over all training and validation tokens. Latents are grouped as early, middle or late according to whether at least half of the squared mass of their temporal profile lies on transition $0\to1$, on transitions $1\to2$ and $2\to3$, or on transitions $3\to4$ and $4\to5$. Within each group, latents active on $5$--$35\%$ of tokens are taken in order of score, skipping any whose per-image activations correlate at $0.8$ or more with a latent already kept, until eight are kept.

Figure~\ref{fig:spatiotemporal-latents} shows the first-ranked early latent and the second-ranked middle latent; the first-ranked middle latent has a spatially diffuse map. The late latent is the only one of the eight kept late latents whose six top samples share one object; the top samples of the others share no object, and the first-ranked late latent's maps are spatially diffuse.

\begin{figure}[h]
    \centering
    \includegraphics[width=\textwidth]{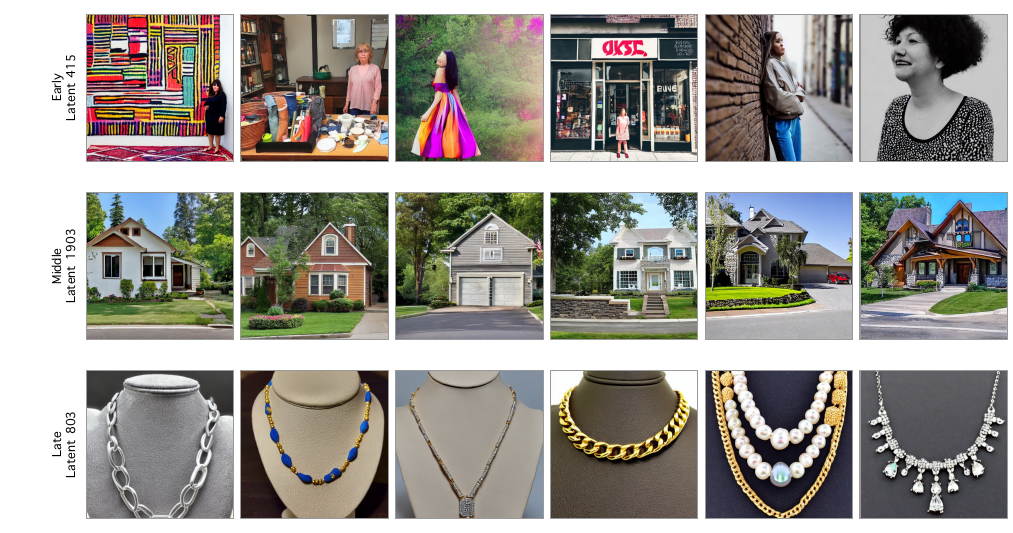}
    \caption{The six top samples of each latent of Figure~\ref{fig:spatiotemporal-latents}. The early latent's all show a woman as the main subject, the middle latent's the front of a house with its lawn or driveway, and the late latent's necklaces on display busts.}
    \label{fig:app-top-samples}
\end{figure}

\paragraph{Maps, self-similarity and top samples.}
A map is the cosine similarity between $\boldsymbol{\phi}_{k,t}(s)$ and the top sample's activation at position $s$ and timestep $t$. The self-similarity between timesteps $t$ and $t'$ is the mean over positions of the cosine similarity between $\boldsymbol{\phi}_{k,t}(s)$ and $\boldsymbol{\phi}_{k,t'}(s)$. Top samples are ranked by the latent's largest activation over tokens, among $10{,}000$ generations from the next $10{,}000$ captions of the same caption stream, after those used for training and validation. Figure~\ref{fig:app-top-samples} shows the six top samples of each latent. Six of the eight late latents kept by the rule put most of their activation on border tokens, but across the dictionary ReSAE's activation is no more concentrated on the border than NoResid+Concat's (Figure~\ref{fig:app-localization}c). The border share is the fraction of a latent's total activation, summed over validation images, that falls on the $28$ border tokens of the $8\times8$ grid; a spatially uniform latent has $28/64=0.44$.

\section{Steering}\label{app:steering}
The dictionaries are trained on a subsampled denoising trajectory, while steering acts at every denoising step. We therefore hold the steering vector of timestep $t$ constant from the step of timestep $t$ up to the step of timestep $t+1$. All steering interventions are applied to the conditional branch of the classifier-free guidance U-Net evaluation; the unconditional branch is left unchanged.

\paragraph{Steering at matched distortion.}
On all $296$ RIEBench pairs, each method steers the source generation toward the target prompt. For a dictionary, the direction is the full difference between the codes of the target and the source generation, each averaged over the edited region of its own generation (the source or target mask of Appendix~\ref{app:onset}) at the six timesteps, decoded and mapped to a feature trajectory. It is added at every token of the source's edited region, with the trajectory of that token's position, using the sampler of Appendix~\ref{app:onset}; the six pairs with an empty mask are left unedited. The strength $\beta\in\{0,0.1875,0.375,0.75,1.5\}$ is the root-mean-square size of the update in units of the activation standard deviation. At each strength we measure the gain in CLIP similarity to the target prompt over the unsteered source image and the LPIPS distance to that image, average both over pairs, and interpolate the gain linearly in mean LPIPS between adjacent strengths to fixed distances of $0.15$ and $0.25$, without extrapolation. Intervals are from $1{,}000$ paired bootstrap resamples of pairs.

Table~\ref{tab:app-steering} compares the trajectory SAEs, the per-timestep SAEs, the Matryoshka SAE, two dictionaries shared across timesteps (one trained on the first timestep only and reused at every timestep, in the manner of \citet{surkovOneStepEnoughSparse2025}, and the Shared Dictionary), and an intervention without a dictionary, which adds at every token of the edited region the difference between the mean activations of the target's and the source's edited regions. Of the eight methods at default sparsity, ReSAE gives the largest gain at both distortion levels. Most paired differences between methods are within their intervals; the one that excludes zero favors ReSAE over the Shared Dictionary at LPIPS $0.25$. At matched density, with $10$ active latents per timestep ($60$ per trajectory, close to ReSAE's $50$), the per-timestep SAEs rank first and second at both distortion levels, but neither differs resolvably from ReSAE (paired differences from $-0.05$ to $-0.01$, with every interval including zero). Steering at matched distortion therefore does not separate the designs, and ReSAE is at or near the top at both distortion levels.

\begin{table}[h]
\caption{Gain in target CLIP similarity at matched LPIPS distance to the source, with 95\% intervals over pairs. The rows with $10$ active latents per timestep are the matched-density per-timestep SAEs. Best in bold, second best underlined.}\label{tab:app-steering}
\centering\small
\begin{tabular}{@{}lcc@{}}
\toprule
method & LPIPS 0.15 & LPIPS 0.25 \\
\midrule
ReSAE & 0.184 [0.058, 0.304] & 0.337 [0.195, 0.489] \\
Resid+NoConcat, $50$ per timestep & 0.163 [0.047, 0.277] & 0.279 [0.113, 0.435] \\
Resid+NoConcat, $10$ per timestep & \textbf{0.198} [0.076, 0.321] & \underline{0.365} [0.198, 0.535] \\
NoResid+Concat & 0.146 [0.014, 0.265] & 0.330 [0.175, 0.487] \\
NoResid+NoConcat, $50$ per timestep & 0.099 [$-$0.024, 0.225] & 0.255 [0.080, 0.419] \\
NoResid+NoConcat, $10$ per timestep & \underline{0.190} [0.072, 0.305] & \textbf{0.387} [0.228, 0.530] \\
Matryoshka & 0.155 [0.023, 0.277] & 0.317 [0.167, 0.469] \\
single-timestep dictionary reused at every timestep & 0.135 [0.010, 0.252] & 0.251 [0.076, 0.413] \\
Shared Dictionary & 0.106 [$-$0.005, 0.221] & 0.204 [0.031, 0.371] \\
no dictionary & 0.139 [0.018, 0.259] & 0.319 [0.157, 0.486] \\
\bottomrule
\end{tabular}
\end{table}

\section{Diffusion Transformer}\label{app:dit}

\paragraph{Setup.}
We use DiT-XL/2 at $256\times256$ resolution with the DDIM sampler (25 steps, guidance scale 1.0) and capture the residual update of the 14th of 28 transformer blocks (its output minus its input), for $256$ patch tokens of dimension $1{,}152$ at the five timesteps $960$, $720$, $480$, $240$ and $0$ of the noise schedule. Dictionaries are trained on $10{,}000$ class-conditional generations and evaluated on $2{,}000$. ReSAE has $2{,}880$ latents with $50$ active latents per trajectory. The TIDE-style SAE shares one dictionary of $13{,}248$ latents across timesteps with a timestep-conditioned encoder \citep{huangTIDETemporalAware2026} and has $10$ active latents per timestep; its size matches ReSAE's parameter count ($33.2$M).

\paragraph{Linear dynamics.}
We fit the linear predictors either per patch position or shared across positions. Position-specific predictors fit the training data better but generalize worse; pooled over transitions their validation explained variance is $0.40$ against $0.43$ for shared predictors (difference $-0.029$ $[-0.030,-0.028]$), with a gap between training and validation of about $0.13$ against $0.01$. The main text uses shared predictors; with position-specific predictors ReSAE explains $0.51$ of the variance of the raw activations.

\paragraph{Class switch.}
For $500$ pairs of ImageNet classes, the class label $y$ is switched to $(y+500)\bmod 1000$ at timestep $720$, $480$ or $240$, giving $1{,}500$ examples and chance $1/3$. The code difference is averaged over all $256$ tokens, since a class switch has no edited region. The classifier and cross-validation are those of Appendix~\ref{app:onset}, grouped by class pair. The five-latent comparison was specified in advance as the primary one.

\begin{table}[h]
\caption{Class-switch onset accuracy on DiT-XL/2 with 95\% intervals (chance $1/3$). Explained variance of reconstruction from the sparse code is $0.59$ for ReSAE and $0.57$ for the TIDE-style SAE on the raw activations, $0.39$ and $0.34$ on the transition residuals, and $0.52$ and $0.55$ under the TIDE-style SAE's own normalized-timestep metric.}\label{tab:app-dit}
\centering\small
\begin{tabular}{@{}lccc@{}}
\toprule
classifier & ReSAE & TIDE-style SAE & difference \\
\midrule
five latents & 0.621 [0.599, 0.643] & 0.541 [0.517, 0.565] & $+0.081$ [$+0.050$, $+0.111$] \\
five latents, unit-normalized & 0.605 [0.582, 0.627] & 0.627 [0.604, 0.650] & $-0.022$ [$-0.053$, $+0.009$] \\
all latents & 0.803 [0.783, 0.822] & 0.509 [0.488, 0.531] & $+0.293$ [$+0.267$, $+0.320$] \\
all latents, unit-normalized & 0.805 [0.784, 0.825] & 0.536 [0.512, 0.560] & $+0.269$ [$+0.240$, $+0.297$] \\
magnitude only & 0.855 [0.835, 0.873] & 0.901 [0.885, 0.917] & $-0.047$ [$-0.067$, $-0.027$] \\
activation difference, raw & \multicolumn{2}{c}{0.635 [0.614, 0.657]} & \\
activation difference, residual & \multicolumn{2}{c}{0.683 [0.660, 0.705]} & \\
\bottomrule
\end{tabular}
\end{table}

\end{document}